\documentclass[journal]{IEEEtran}
\IEEEoverridecommandlockouts
\usepackage{cite}
\usepackage{amsmath,amssymb,amsfonts}
\usepackage{graphicx}
\usepackage{textcomp}

\usepackage[utf8]{inputenc}
\inputencoding{latin1}
\inputencoding{utf8}
\usepackage{microtype}
\usepackage{lipsum}
\usepackage{graphicx}
\usepackage[dvipsnames]{xcolor}
\usepackage{booktabs} 
\usepackage{amssymb,amsthm,amsmath}
\usepackage{bbm}
\usepackage{float}
\usepackage[misc]{ifsym}
\usepackage{listings,newtxtt}
\usepackage{algorithmicx,algorithm,eqparbox,array}
\algnewcommand{\LineComment}[1]{\State // #1}
\usepackage[noend]{algpseudocode}
\usepackage{tcolorbox}
\usepackage{hyperref}
\usepackage{etoolbox}
\usepackage{tikz}
\usetikzlibrary{shapes}
\usetikzlibrary{calc}

\makeatletter
\apptocmd\@makecaption{\par}{}{%
  \errmessage{\noexpand\@makecaption could not be patched}%
}
\makeatother

\usepackage{environ}
\makeatletter
\newsavebox{\measure@tikzpicture}
\NewEnviron{scaletikzpicturetowidth}[1]{%
  \def\tikz@width{#1}%
  \def\tikzscale{1}\begin{lrbox}{\measure@tikzpicture}%
  \BODY
  \end{lrbox}%
  \pgfmathparse{#1/\wd\measure@tikzpicture}%
  \edef\tikzscale{\pgfmathresult}%
  \BODY
}
\makeatother

\colorlet{CommentColor}{MidnightBlue}
\colorlet{KeywordColor}{Fuchsia}
\colorlet{LiteralColor}{Mahogany}
\colorlet{KeywordColor}{Black}

\makeatletter
\def\mathcolor#1#{\@mathcolor{#1}}
\def\@mathcolor#1#2#3{%
  \protect\leavevmode
  \begingroup
    \color#1{#2}#3%
  \endgroup
}
\makeatother

\algnewcommand{\LineComment}[1]{\State // #1}

\theoremstyle{plain}
\newtheorem{thm}{Theorem} 
\theoremstyle{definition}
\newtheorem{defn}[thm]{Definition} 

\begin{document}

\title{Typed Graph Networks}
  \author{
      Marcelo~O.R.~Prates*,
      Pedro~H.C.~Avelar*,
      Henrique~Lemos*,
      Marco~Gori,~\IEEEmembership{Fellow,~IEEE,}
      and~Luis~Lamb,~\IEEEmembership{Member,~IEEE}

  \thanks{This work is partly supported by Coordenação de Aperfeiçoamento de Pessoal de Nível Superior (CAPES) - Finance Code 001 and by the Brazilian Research Council CNPq.}
  \thanks{M.O.R. Prates, P.H.C. Avelar, H. Lemos and L. Lamb are with the Institute of Informatics, Federal University of Rio Grande do Sul, Porto Alegre, Brazil (e-mail: \{morprates,phcavelar,hlsantos,lamb\}@inf.ufrgs.br).}
  \thanks{M. Gori is with the Department of Information Engineering and Mathematical Sciences, University of Siena, Siena, Italy (e-mail: marco@dii.unisi.it).}
  \thanks{* Equal contribution}}


\markboth{IEEE Transactions on Neural Networks and Learning Systems, VOL.~?, NO.~?, ?~?
}{Avelar \MakeLowercase{\textit{et al.}}: Typed Graph Networks}

\maketitle

\begin{abstract}
    Recently, the deep learning community has given growing attention to neural architectures engineered to learn problems in relational domains. Convolutional Neural Networks employ parameter sharing over the image domain, tying the weights of neural connections on a grid topology and thus enforcing the learning of a number of convolutional kernels. By instantiating trainable neural modules and assembling them in varied configurations (apart from grids), one can enforce parameter sharing over graphs, yielding models which can effectively be fed with relational data. In this context, vertices in a graph can be projected into a hyperdimensional real space and iteratively refined over many message-passing iterations in an end-to-end differentiable architecture. Architectures of this family have been referred to with several definitions in the literature, such as Graph Neural Networks, Message-passing Neural Networks, Relational Networks and Graph Networks. In this paper, we revisit the original Graph Neural Network model and show that it generalises many of the recent models, which in turn benefit from the insight of thinking about vertex \textbf{types}. To illustrate the generality of the original model, we present a Graph Neural Network formalisation, which partitions the vertices of a graph into a number of types. Each type represents an entity in the ontology of the problem one wants to learn. This allows -  for instance - one to assign embeddings to edges, hyperedges, and any number of global attributes of the graph. As a companion to this paper we provide a Python/Tensorflow library to facilitate the development of such architectures, with which we instantiate the formalisation to reproduce a number of models proposed in the current literature.
\end{abstract}

\section{Introduction}

Machine learning in general and deep learning (DL) in particular have sported significant advances in the last decade. Deep convolutional networks have consistently pushed the state-of-the-art in image classification \cite{simonyan2014very,he2016deep}; neural implementations of the $Q$-function \cite{watkins1992q} allowed for effectively training reinforcement learning agents on huge combinatorial state spaces such as the pixels of Atari games \cite{mnih2013playing,mnih2015human} and the marriage between symbolic tree search and DL evaluation functions has yielded the mastering of the games of Go and chess, a longstanding testbed for artificial intelligence research \cite{silver2016mastering,silver2017masteringgo,silver2017masteringchessshogi}. Machine learning has been successful against human champions by training itself to superhuman levels in Chess and Shogi solely from self-play \cite{silver2017masteringchessshogi}. Insights from game theory have allowed DL models to transcend classification tasks and produce generative inputs, with Generative Adversarial Neural Networks rapidly nearing photo-realistic results \cite{karras2017progressive}. With an ever increasing array of diverse complex scenarios being successfully projected onto the continuous landscapes of DL models' parameter spaces, a new and fundamental frontier in AI research leads to the marriage between DL and the discrete, relational realms up until very recently reserved to the long-parted branch of symbolic AI. Empowering deep learning to tackle combinatorial generalisation is now seen as a key path forward in AI research \cite{battaglia2018relational}. 

The marriage between machine learning and combinatorial optimisation is seen with optimism, as machine learning has shown significant promise in replacing expert knowledge -- nowadays a key component of operations research -- for a wide range of complex domains \cite{bengio2018machine}. Machine learning is suited for problem landscapes with no simple mathematical structure, a feature sported by hard combinatorial problems. In this context, the barriers often imposed by the data hunger of ML models is minimised by the fact that exact solutions are available for combinatorial problems, which enables one to circumvent the problem by algorithmically generating labelled training datasets. It is also expected that ML models will be able to decompose combinatorial problems into smaller and simpler learning tasks \cite{bengio2018machine}.

Strong contenders for bridging deep learning with combinatorial optimisation are neural network models that work on \emph{relational} data, i.e. neural network (NN) models which exploit invariants in combinatorial domains by implementing parameter sharing over graphs. These networks which work on relational data can be seen as generalised variants of convolutional networks where operations analogous to convolutions are not necessarily performed over grids. Concretely, such models can be implemented by instantiating neural modules, which, in the same way as convolutional kernels, will be repeatedly applied to the entire problem. While convolutional kernels sweep over rectangular pixel windows, however, in this context we sweep over nodes in an arbitrary graph. This family of architectures \cite{gori2005new,scarselli2009graph,li2015gated,gilmer2017neural} has produced significant results in the last years on a wide range of applications such as ranking web pages \cite{scarselli2005graph}, visual scene understanding \cite{raposo2017discovering,santoro2017simple}, relational reasoning over the dynamics of physical systems \cite{battaglia2016interaction}, learning on knowledge graphs \cite{bordes2013translating} and predicting quantum chemistry properties \cite{gilmer2017neural}.

\begin{figure}[t]
    \centering
    \includegraphics[width=\linewidth]{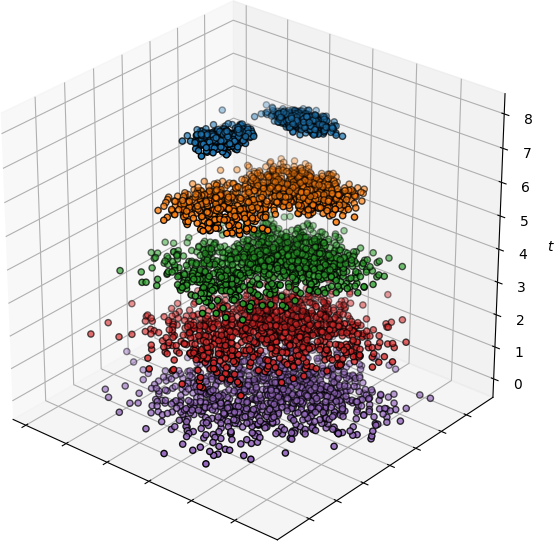}
    \caption{The computation performed by a Graph Neural Network can be interpreted as the iterative refinement (over many message-passing iterations) of an initial projection $\mathcal{P}_0: \mathcal{V} \rightarrow \mathbb{R}^d$ of graph vertices into hyperdimensional space. The options for choosing the initial projection $\mathcal{P}_0$ are manyfold: it can be computed as a function of vertex attributes for a vertex-labelled graph, it can map all vertices to the same point $\in \mathbb{R}^d$ or, as this Figure shows, vertex projections can be chosen at random. A successfully trained GNN model will be capable of refining a projection which captures some property of the learned problem, for example a 2-partitioning of $\mathcal{V}$. The Figure shows a pictorial representation of a sequence of progressively refined projections $\mathcal{P}_0, \mathcal{P}_2, \mathcal{P}_4, \mathcal{P}_6, \mathcal{P}_8$ over $t_{max} = 8$ message-passing timesteps.}
    \label{fig:GNN-viz}
\end{figure}

Architectures in this family project each vertex in the input graph into a hyperdimensional real space, assigning a vector $\in \mathbb{R}^d$ for each vertex. This projection is iteratively refined (Figure \ref{fig:GNN-viz}) as vertices receive messages along their incoming edges and use them as inputs to a function which updates their current embedding. In this context, the sole trainable components of such a model are a function $\mu$ which computes a message from a vertex embedding and a function $\phi$ which updates a vertex embedding given a set of messages. Over many iterations of message-passing, one should expect that vertices become enriched with information about their neighbourhood, ultimately lending themselves to some kind of treatment which allows a complex function on graphs to be learned via gradient descent.

The remainder of this paper is structured as follows: Section \ref{sec:rel-work} focuses on describing a family of neural models which were all derived from the original Graph Neural Network model (GNN) \cite{scarselli2009graph}, we specifically lay emphasis on how the Graph Network model, proposed by \cite{battaglia2018relational}, did not bring to light additional expressiveness regarding the original GNN model -  although their work played an important role on surveying these family of neural models and discussing how these models can benefit from relational inductive bias towards relational learning. In Section \ref{sec:form-tgn} we present a new formalisation to the GNN model, also proposing a new terminology, which emphasises its generalisation capability. Throughout Section \ref{sec:lib-tgn} we briefly describe and showcase our open-source Python library, which we believe allows an efficient and intelligible implementation within a GNN framework. Finally, in Section \ref{sec:conc} we present our final thoughts on why it was important to revisit the GNN model and to highlight its generality over the recent models.

\section{Related Work}
\label{sec:rel-work}

The family of neural models discussed in this paper (or variants thereof) is referred to with different names in the scientific literature, such as message-passing neural networks \cite{gilmer2017neural}, recurrent relational networks \cite{palm2017recurrent} and graph neural networks \cite{scarselli2009graph}. With deep learning models which exploit relational structure sporting increasing popularity in the last years, the multitude of definitions can be confusing and the necessity of unifying most of them into a single formalisation becomes paramount. 
In this context, \cite{battaglia2018relational} have proposed a ``Graph Network'' model whose goal is to ``generalise and extend several lines of work'' in this area. The Graph Network formalisation includes \textbf{nodes}, \textbf{edges} and the \textbf{graph} in its ontology, which is to say that the model projects each node, edge and additionally the entire graph on an hyper-dimensional space. These projections (also called embeddings) may be iteratively refined over many iterations of message-passing in which nodes communicate with edges (and vice-versa) and both of them communicate with a global embedding representing the entire graph.
\begin{algorithm}[t]
\caption{Graph Network Formalisation \cite{battaglia2018relational} ran for $t$ message-passing timesteps.}\label{alg:GN}
\begin{algorithmic}[1]\small
\Procedure{GraphNetwork}{$\mathcal{G} = (\mathcal{V}, \mathcal{E}, \mathbf{u})$}

\For{$t = 1 \dots t_{max}$}
    \For{$k = 1 \dots |\mathcal{E}|$}
        \LineComment{Compute updated edge embeddings}
        \State $\mathbf{{e}_k}^{(t+1)} \leftarrow \phi_{e}(\mathbf{e}_k^{(t)}, \mathbf{v}_{r_k}^{(t)}, \mathbf{v}_{s_k}^{(t)}, \mathbf{u}^{(t)})$
    \EndFor
    \For{$i = 1 \dots |\mathcal{V}|$}
        \LineComment{Aggregate messages sent to node $i$ into $\overline{\mu}$}
        \State $\overline{\mu} \leftarrow \rho_{e \rightarrow v}(\{ \mathbf{e}_k^{(t+1)} ~|~ r_k=i \})$
        \LineComment{Update vertex embedding}
        \State $\mathbf{v}_i^{(t+1)} \leftarrow \phi_{v}(\mathbf{v}_i^{(t)},\overline{\mu},\mathbf{u}^{(t)})$
    \EndFor
    \LineComment{Compute updated global embedding}
    \State $\mathbf{u}^{(t+1)} \leftarrow \phi_{u}(\mathbf{u},\{\mathbf{v}_i^{(t+1)}\}_{i=1 \dots |\mathcal{V}|},\{\mathbf{e}_i^{(t+1)}\}_{i=1 \dots |\mathcal{E}|})$
\EndFor
\LineComment{Return refined vertex embeddings, edge embeddings and global embedding}
\State \Return $\{\mathbf{v}_i^{(t_{max})}\}_{i=1 \dots |\mathcal{V}|}, \{\mathbf{e}_i^{(t_{max})}\}_{i=1 \dots |\mathcal{E}|}, \mathbf{u}^{(t_{max})}$
\EndProcedure
\end{algorithmic}
\end{algorithm}
The Graph Network formalisation is especially appealing for graph-related applications in which both nodes and edges can have labels associated to them and in which accumulating local information into global attributes can make sense (for example the global clustering coefficient of a graph \cite{watts1998collective}).
In other cases one very much would like a formalisation which naturally supports hypergraph descriptions.
Consider the problem of boolean satisfiability, for example, for which an effective solution has been learned with an architecture of the Graph Neural Network family \cite{selsam2018learning}. In the ontology of SAT there are \emph{literals} (i.e. possibly negated boolean variables such as $x_3$ or $\neg x_7$) and \emph{clauses} (disjunctions of literals such as $(x_1 \vee x_3 \vee \neg x_2)$). A SAT instance can be described by the set of its literals $\{x_1, \neg x_1, x_2, \neg x_2 \dots x_n, \neg x_n\}$ and by the set of its clauses, each of which can be described as a set of literals (for example $\{x_1, x_3, \neg x_2\}$). In this context, each SAT instance can be seen as an hypergraph in which literals correspond to nodes and clauses correspond to hyperedges connecting a given number of nodes. 
This paper provides a formalisation to support also hypergraph descriptions, that are defined in a simple and elegant way.

\section{A Formalisation of Typed Graph Networks}
\label{sec:form-tgn}

An interesting feature of the graph neural network model early introduced in~\cite{scarselli2009graph} is that any node $k_n$ can be of a given type, so as it can be associated with its own transition function, output function, and parameters set. Apart from this remark, however, there was no mention in the paper on the way this feature can play a crucial role in the appropriate modelling of cognitive tasks.
Clearly, the adoption of typed nodes allow us to compute multiple global properties of an input graph. In this context, it would be useful to have not one but $k$ embeddings for global attributes, along with corresponding update functions $\phi_{u_1}, \phi_{u_2} \dots \phi_{u_k}$. 
Thus posed, we wish to emphasise the original authors' contribution by proposing an updated terminology focused on the possibility of having several types of objects, which is in line with more recent work \cite{gilmer2017neural} and improves understandability, removing unnecessary complexities from the original model such as the support for multiple edge types -- since the concept of node types immediately adds the same expressiveness. 
The formalisation of the Typed Graph Network (TGN) proposed in this paper allows us to face successfully a number of challenging graph problems and, moreover, it opens the doors to the massive adoption of TGN to other problems, thanks to a related TGN Python library, that is made available.

\begin{algorithm*}[t]
\caption{Typed Graph Network Model}\label{alg:TGN}
\begin{algorithmic}[1]\small
\Procedure{TGN}{$\mathcal{G} = (\mathcal{V} = \bigcup\limits_{i=1}^{N}{\mathcal{V}_i}, \mathcal{E} = \bigcup\limits_{k=1}^{K}{{\mathcal{E}_k}})$} \Comment{The input for a TGN is a graph whose vertices are partitioned into $N$ \textbf{vertex types}, and the edges into $K$ \textbf{edge types}}

\For{$k=1 \dots K$}
    \State $\mathbf{M}_{k}[a,b] = \mathbbm{1} \{ (v_a,v_b) \in {\mathcal{E}_k}\}$ \Comment{Compute an adjacency matrix between types $\#i$ and $\#j$}
\EndFor

\For{$i=1 \dots n$}
    \State Init vertex embeddings $\mathbf{V_i}^{(1)}[a] \in \tau_i ~|~ \forall v_a \in \mathcal{V}_i$
\EndFor

\For{$t=1 \dots t_{max}$} \Comment{Run for $t_{max}$ message-passing iterations}
    \For{$i = 1 \dots N$} \Comment{For each receiving type $\#i$}
        \For{$\mu_k \in \mathcal{M} ~|~ \mu_k : \tau_i \rightarrow \tau_j $} \Comment{For each message sent from type $\#j$}
            \State $\overline{\mu}_k \leftarrow \mathbf{M}_{k} \times \mu(\mathbf{V_j}^{(t)})$ \Comment{Accumulate messages sent to vertices of type $\#i$ by vertices of type $\#j$}
        \EndFor
        \State $\mathbf{V_i}^{(t+1)} \leftarrow \phi_{i}(\mathbf{V_i}^{(t)}, \{ \overline{\mu}_k ~|~ \mu_k \in \mathcal{M} , \mu_k : \tau_i \rightarrow \tau_j \})$ \Comment{Compute updated embeddings for type $\#i$}
    \EndFor
\EndFor
\State \Return $\{\mathbf{V_i}^{(t_{max})} ~|~ i=1 \dots n\}$ \Comment{Return set of refined embeddings over $t_{max}$ iterations}
\EndProcedure
\end{algorithmic}
\end{algorithm*}

In the Graph Network model, node, edge and graph embeddings also differ by the way they are connected. Edge embeddings connect to the node embeddings corresponding to their source and target vertices (i.e. the embedding for edge $(i,j)$ is connected to the embeddings for node $i$ and $j$) and the graph embedding connects to all edge and all node embeddings. If we once again think about this description by identifying node, edge and graph embeddings as just node embeddings of different \textbf{types}, all these connections can be described by an $|\mathcal{E}| \times |\mathcal{V}|$ adjacency matrix mapping edges to source vertices, an $|\mathcal{E}| \times |\mathcal{V}|$ adjacency matrix mapping edges to target vertices, a (complete) adjacency matrix $|\mathcal{E}| \times |\mathcal{G}|$ mapping (all) edges to the graph embedding and a (complete) adjacency matrix $|\mathcal{V}| \times |\mathcal{G}|$ mapping (all) nodes to the graph embedding. In this context, it is straightforward to see how the issue with hyperedges can be resolved: hyperedges $\in \mathcal{H}$ can be associated with vertices $\in \mathcal{V}$ by an adjacency matrix $|\mathcal{H}| \times |\mathcal{V}|$. Nothing changes apart from the fact that we drop the constraint that all rows in the adjacency matrices between edges and vertices must add up to unity (i.e. each hyperedge can connect to possibly more than one vertex).

We can see that by replacing the hard-coded connections of the Graph Network model with connections between objects of different types we allow for more versatility, enabling to train models on domains with richer structure than regular graphs. In many applications, the edge and graph embeddings become appendages in the Graph Network model. By contrast, in the context of the Typed Graph Network model we will define shortly, we only instantiate these objects when they are actually present in the ontology of the problem we want to learn.

\begin{tcolorbox}
\begin{defn}[Typed Graph Network] \label{def:TGN}
    A Typed Graph Network with $N$ types is described by
    \begin{enumerate}
        \item A set of \textbf{embedding sizes} $n_1, n_2 \dots n_N$
        \item A set of \textbf{types} $\mathcal{T} = \{ \tau_i \in \mathbb{R}^{n_i} ~|~ i=1 \dots N \}$.
        \item A set of $K$ \textbf{message} functions \\${\mathcal{M} = \{\mu: \tau_1 \rightarrow \tau_2 ~|~ \tau_1,\tau_2 \in \mathcal{T}\}}$ \\ OBS. Note that for each type combination $(\tau_1,\tau_2)$ one can define many message functions.
        \item And a set of \textbf{update} functions \\${\phi_{i}: \mathbb{R}^{n_i + D(i)} \rightarrow \mathbb{R}^{n_i} ~|~ \forall i \in \{1 \dots N\}}$, where $D(i) = \displaystyle\sum_{\mu: \tau_j \rightarrow \tau_i \in \mathcal{M}}{n_i}$
    \end{enumerate}
    Where the message functions $\tau_{i \rightarrow j}$ and the update functions $\phi_i$ are the sole trainable components of the model.
\end{defn}
\end{tcolorbox}

\begin{figure}[h]
    \centering
    \begin{scaletikzpicturetowidth}{.9\linewidth}
    \begin{tikzpicture}[auto, node distance=2.2cm, scale=\tikzscale, every node/.style={transform shape}]
        \node [draw, cloud, cloud puffs=24.7, cloud ignores aspect, fill=GreenYellow!40!white!60, name=inneighbourhood, align=center] {Time $(t)$\\Incoming\\Neighbourhood};
        \node [draw, rectangle, rounded corners=0.1cm, fill=Red!40!white!60, below right of=inneighbourhood, name=inmsg, align=center] {Message ($\mu$)};
        \node [draw, rectangle, rounded corners=0.1cm, fill=Cyan!40!white!60, below right of=inmsg, name=update, align=center] {Time $(t)$ \\ Update ($\phi$)};
        \node [draw, rectangle, rounded corners=0.1cm, fill=Cyan!40!white!60, left of=update, name=updatebefore, align=center] {Time $(t-1)$ \\ Update ($\phi$)};
        \node [left of=updatebefore, name=updatebefore2] {};
        \node [draw, rectangle, rounded corners=0.1cm, fill=Cyan!40!white!60, right of=update, name=updateafter, align=center] {Time $(t+1)$ \\ Update ($\phi$)};
        \node [right of=updateafter, name=updateafter2] {};
        \node [draw, rectangle, rounded corners=0.1cm, fill=Red!40!white!60, below right of=update, name=outmsg, align=center] {Message ($\mu$)};
        \node [draw, cloud, cloud puffs=24.7, below right of=outmsg, cloud ignores aspect, fill=Emerald!20!white!80, name=outneighbourhood, align=center] {Time $(t+1)$\\Outgoing\\Neighbourhood};
        
        \draw[->,>=stealth] ($(inneighbourhood) + (0cm,1.5cm)$) -- node[auto,sloped,font=\footnotesize] {Time} ($(inneighbourhood) + (7cm,1.5cm)$);
    
        \path[every node/.style={font=\sffamily\small}]
        (inneighbourhood)
            edge[->]
                node [] {} (inmsg)
        (inmsg)
            edge[->] node [] {}
                (update)
        (updatebefore2)
            edge[->, dashed, bend right=80] node [below] {}
                (updatebefore)
        (updatebefore)
            edge[->, dashed, bend right=80] node [below] {}
                (update)
        (update)
            edge[->] node [] {}
                (outmsg)
        (update)
            edge[->, dashed, bend right=80] node [below] {}
                (updateafter)
        (updateafter)
            edge[->, dashed, bend right=80] node [below] {}
                (updateafter2)
        (outmsg)
            edge[->]  node [] {}
                (outneighbourhood)
        ;
    \end{tikzpicture}
    \end{scaletikzpicturetowidth}
    \caption{Pictorial representation of a Typed Graph Network from the perspective of a vertex $v$. A set of embeddings is received from vertices in its \colorbox{GreenYellow!60!white!40}{incoming neighbourhood}, a message is computed from each embedding with the message function \colorbox{red!40!white!60}{$\mu$} and messages are aggregated and fed to the update function \colorbox{Cyan!40!white!60}{$\phi$}, which produces an updated embedding for $v$. Simultaneously, $v$ sends messages to vertices in its \colorbox{Emerald!30!white!70}{outgoing neighbourhood}, which will undergo the same update process.}
    \label{fig:nodeview}
\end{figure}
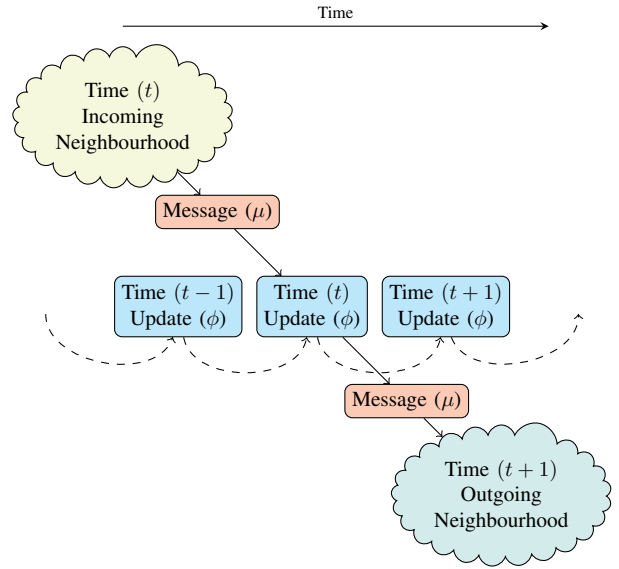

The fundamental insight behind Typed Graph Networks is that we can project different types of vertices into different real hyperspaces (possibly of differing dimensionalities) and make vertices of different types communicate with one another simply by computing messages from one projection to the other. In this context, a message to be computed from projections $\in \tau_i$ to projections $\in \tau_j$ is implemented by a (trainable, parameterised) function $\mu : \tau_i \rightarrow \tau_j$.

Algorithm \ref{alg:TGN} describes the Typed Graph Network model in detail. The TGN model may run for many message-passing iterations (Line 7). The step performed at a single iteration is explained here. For each type $i$ (Line 8) and for each message function $\mu_k : \tau_i \rightarrow \tau_j$ (Line 9) we accumulate all embeddings sent to vertices $v_a \in \mathcal{V}_i$ by vertices $v_b \in \mathcal{V}_j$, computing a message from each of them with $\mu_k$. Notice that this accumulation can be done through matrix multiplication, as each line $\overline{\mu}_k[a,:]$ of the matrix multiplication $\overline{\mu}_k = \mathbf{M}_{k} \times \mu(\mathbf{V_j}^{(t)})$ is computed as

\begin{equation}
    \overline{\mu}_k[a,:] = \sum_{v_b \in \mathcal{V}_j}{\mathbf{M}_{k}[a,b] \times \mu_k(\mathbf{V_j}^{(t)}[b])}
\end{equation}

In other words, $\mu_k[a,:]$ corresponds to the element-wise sum of all messages received by vertex $v_a \in \mathcal{V}_i$ from vertices $v_b \in \mathcal{V}_j$. Having computed aggregated message tensors $\overline{\mu}_k$ for all message functions ${\mu_k \in \mathbf{M} ~|~ \mu_k : \tau_i \rightarrow \tau_j}$, we can now update vertex embeddings of type $i$ as 

\begin{equation}
\mathbf{V_i}^{(t+1)} \leftarrow \phi_i(\mathbf{V_i}^{(t)}, \{ \overline{\mu}_k ~|~ \mu_k \in \mathcal{M} , \mu_k : \tau_i \rightarrow \tau_j \})
\end{equation}

i.e., we update vertex embeddings of type $i$ by feeding the update function with the current vertex embeddings $\mathbf{V_i}^{(t)}$ and the aggregated message tensors $\{ \overline{\mu}_k ~|~ \mu_k \in \mathcal{M} , \mu_k : \tau_i \rightarrow \tau_j \}$ corresponding to all messages received from types $\in \mathcal{T}$ (note that a type can send messages to nodes of the same type as itself i.e. $\mu : \tau_i \rightarrow \tau_i$). Note that here we fix the messages to be aggregated with sums due to the convenience of matrix multiplication, but one could choose whatever aggregation function it prefers.

\section{A Library for Typed Graph Networks}
\label{sec:lib-tgn}

As a companion to this paper, we provide a Python/Tensorflow library to ease the definition of TGN models\footnote{Available at \url{https://github.com/machine-reasoning-ufrgs/typed-graph-network}}. Our library allows one to specify a TGN succinctly, at a description level similar to that of the formalisation in Algorithm \ref{alg:TGN}. This description is compiled into a set of Parameters that, when the model is called upon a input, constructs a computation graph accordingly, thus yielding a module whose inputs and outputs can be connected to any other operations at the desired point in the trainable model's pipeline. It is important to note here that our implementation, due to practical reasons, implements the update function exclusively as a LSTM-like RNN, which operates on both the last hidden state and output, the message functions as Multilayer Perceptrons and the aggregation of messages as a sum aggregation.

The TGN builder identifies each vertex type by a string, for example {\color{LiteralColor}'V'} for vertex vertices and {\color{LiteralColor}'E'} for edge vertices. Adjacency matrices are also identified by strings, for example {\color{LiteralColor}'EV'} for an edge-to-vertex adjacency matrix $\in \mathbb{B}^{|\mathcal{E}| \times |\mathcal{V}|}$, as well as message functions, for example {\color{LiteralColor}'V\_cast\_E'} for a message function $\tau_{\mathcal{V} \rightarrow \mathcal{E}}: \mathbb{R}^{n_{\mathcal{V}}} \rightarrow \mathbb{R}^{n_{\mathcal{E}}}$ mapping vertex embeddings to edge embeddings. An update function is automatically instantiated for each type, but each of its arguments must be specified by an adjacency matrix, a message function and the type of the sender vertices.

Concretely, the TGN builder receives 4 arguments:
\begin{enumerate}
    \item A Python dictionary mapping type names to embedding sizes, such as \{{\color{LiteralColor}'V'}: $d_v$, {\color{LiteralColor}'E'}: $d_e$\}. Equivalent to $\tau_{\mathcal{V}} = \mathbb{R}^{d_v}, \tau_{\mathcal{E}} = \mathbb{R}^{d_e}$.
    \item A Python dictionary mapping matrix names to 2-uples of type names, such as \{{\color{LiteralColor}'EV'}: ({\color{LiteralColor}'E'},{\color{LiteralColor}'V'})\}. Equivalent to: $\mathbf{EV} \in \mathbb{R}^{|\mathcal{E}| \times |\mathcal{V}|}$.
    \item A Python dictionary mapping message function names to 2-uples of type names, such as\\ \{{\color{LiteralColor}'V\_cast\_E}: ({\color{LiteralColor}'V'},{\color{LiteralColor}'E'}), {\color{LiteralColor}'E\_cast\_V'}: ({\color{LiteralColor}'E'},{\color{LiteralColor}'V'})\}. Equivalent to: $\mu_{\mathcal{V} \rightarrow \mathcal{E}}: \tau_{\mathcal{V}} \rightarrow \tau_{\mathcal{E}}, \mu_{\mathcal{E} \rightarrow \mathcal{V}}: \tau_{\mathcal{E}} \rightarrow \tau_{\mathcal{V}}$.
    \item A Python dictionary mapping type names to lists of Python dictionaries each specifying an aggregation of messages, such as \{{\color{LiteralColor}'E'}: [\{{\color{LiteralColor}'mat'}: {\color{LiteralColor}'EV'}, {\color{LiteralColor}'msg'}: {\color{LiteralColor}'V\_cast\_E'}, {\color{LiteralColor}'var'}: {\color{LiteralColor}'V'}\}]\}. Equivalent to: \\$\mathbf{E}^{(t+1)}, \mathbf{E}_h^{(t+1)} \leftarrow \phi_{\mathcal{E}}(\mathbf{E}_h^{(t)}, \mathbf{M_{\mathcal{E} \mathcal{V}}} \times \mu_{\mathcal{V} \rightarrow \mathcal{E}}(\mathbf{V}^{(t)}))$). Note that this argument corresponds to a \textbf{list} of dictionaries. This is so because the update function can receive multiple arguments.
\end{enumerate}

Then, the TGN itself is called with the desired inputs, effectively coupling it with the rest of the model's pipeline and producing the output of the last states of each type of vertex. This also permits, in case of the Tensorflow implementation, to set the builder to a specific variable scope and choose whether to make use of parameter sharing or not. When coupling the TGN with the rest of the model, it receives 3 dictionaries and 1 integer, with one of the dictionaries being optional, specifying the adjacency matrices, the initial embeddings and the initial hidden-states of the RNNs (being that the hidden state is assumed to be a zero tensor, if missing) as well as how many time-steps of computation should be performed.

Concretely, upon calling the TGN and coupling it with the rest of the pipeline, it receives:

\begin{enumerate}
    \item A Python dictionary mapping matrix names to adjacency matrices between two vertex types, such as \{{\color{LiteralColor}'EV'}: $M_ {|E|\times|V|}\}$.
    \item A Python dictionary mapping type names to the initial embeddings of each vertex of that type, such as \{{\color{LiteralColor}'V'}: $V$, {\color{LiteralColor}'E'}: $E$\}, with $V \in \mathbb{R}^{|V| \times d_v}$ and $E \in \mathbb{R}^{|E| \times d_e}$ being the tensors containing the initial embedding for each of the vertices in the Graph.
    \item A single integer $t_{max}$, defining how many timesteps of message-passing are to be performed.
    \item A Python dictionary mapping type names to the initial hidden states embeddings of each vertex of that type, such as \{{\color{LiteralColor}'V'}: $\mathbf{V}_0$, {\color{LiteralColor}'E'}: $\mathbf{E}_0$\}, with $\mathbf{V}_0 \in \mathbb{R}^{|\mathcal{V}| \times d_v}$ and $\mathbf{E}_0 \in \mathbb{R}^{|\mathcal{E}| \times d_e}$ being the tensors containing the initial hidden-state embedding for each of the vertices in the Graph.
\end{enumerate}

Another, implementation-specific, note that can be raised is also that the batching of different graphs can be done by simply concatenating different graphs, without making any adjacencies between one another. In this way, information from a node in a graph will never reach information in a node in another graph, and thus one can make even better use of parallelism to perform faster computation on the inputs.

On the following subsections, we will give some examples of how to implement some models using the TGN formalisation and our library, and specifically in Subsections~\ref{ssec:neurosat-impl} and \ref{ssec:coloring-impl} we explain how to implement models in which the TGN formalisation works more naturally than others explained in the literature, as explained in \ref{sec:form-tgn}.

\subsection{NeuroSAT}\label{ssec:neurosat-impl}

\cite{selsam2018learning} have shown that neural models in the GNN family can learn to solve the problem of boolean satisfiability with up to $85\%$ accuracy upon being fed with complementary (SAT/UNSAT) instances which differ by a single literal's polarity on a single clause. The insight behind the authors' architecture is to assign embeddings both to literals (i.e. $x_1, \neg x_5$, etc.) and clauses (i.e. $(x_2 \vee \neg x_10 \vee \neg x_3)$). Literals send messages to clauses to which they pertain, and clauses send messages to literals they contain. Additionally, literals send messages to their negated variants (i.e. $x_1$ sends messages to $\neg x_1$ and vice-versa). This can be formalised by the following update equations:

\begin{equation}
\begin{aligned}
    \mathbf{L}^{(t+1)}, \mathbf{L_h}^{(t+1)} \leftarrow \phi_{\mathcal{L}}(\mathbf{L_h}^{(t)}, \mathbf{M_{\mathcal{LC}}}^{\intercal} \times \mu_{\mathcal{C} \rightarrow \mathcal{L}}(\mathbf{C}^{(t)}), \\ \mathbf{M_{\mathcal{LL}}} \times \mathbf{L}^{(t)})) \\\\
    \mathbf{C}^{(t+1)}, \mathbf{C_h}^{(t+1)} \leftarrow \phi_{\mathcal{C}}(\mathbf{C_h}^{(t)}, \\\mathbf{M_{\mathcal{LC}}} \times \mu_{\mathcal{L} \rightarrow \mathcal{C}}(\mathbf{L}^{(t)}))
\end{aligned}
\end{equation}

Where $\mathcal{L} = \{x_1,\neg x_1 \dots x_i, \neg x_i \dots x_N, \neg x_N\}$ is the set of all $2N$ literals, $\mathcal{C} \in (2^{\mathcal{L}})^{M}$\footnote{$2^{\mathbf{A}}$ denotes the \emph{power set} of $\mathbf{A}$} is the set of all $M$ clauses and a CNF formula can be described by an adjacency matrix $\mathbf{M_{\mathcal{LC}}} \in \mathbb{B}^{2N \times M}$ between literals and clauses as ${\mathbf{M_{\mathcal{L} \times \mathcal{C}}} = \mathbbm{1}\{ (l,c) \in \mathcal{L} \times \mathcal{C} ~|~ l \in c \}}$. To connect literals with their negated variants, it suffices to define an adjacency matrix $\mathbf{M_{\mathcal{LL}}} \in \mathbb{B}^{2L \times 2L}$ between literals and literals as ${M_{\mathcal{LL}} = \mathbbm{1}\{ (l_1,l_2) \in \mathcal{L}^2 ~|~ l_1 = \neg l_2 \}}$.

\begin{figure}[h]
    \centering
    \includegraphics[width=\linewidth]{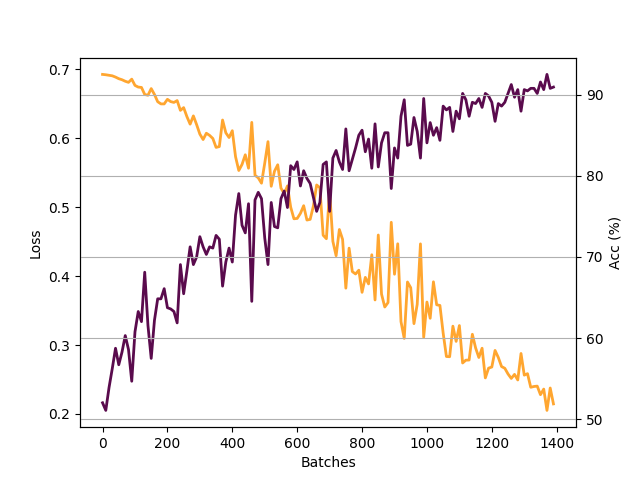}
    \caption{Results obtained with the model proposed by \cite{selsam2018learning} implemented in the Typed Graph Networks library (see Code Listing \ref{TGN-neurosat}).}
    \label{fig:log-neurosat}
\end{figure}

\begin{figure}[h]
\begin{minipage}{\linewidth}
\begin{lstlisting}[language=Python,mathescape=true, caption={TGN kernel of an end-to-end differentiable model to solve the boolean satisfiability problem (SAT), implemented with our Python library. Lines of code are commented with the corresponding equations in the proposed formalisation for TGNs in Algorithm \ref{alg:TGN}.},label=TGN-neurosat]
tgn = TGN(
  {
    # $\mathcolor{CommentColor}{\tau_{\mathcal{L}} = \mathbb{R}^{d_l}}$
    'L': dl,
    # $\mathcolor{CommentColor}{\tau_{\mathcal{C}} = \mathbb{R}^{d_c}}$
    'C': dc
  },
  {
    # $\mathcolor{CommentColor}{\mathbf{M_{\mathcal{LL}}} \in \mathbb{B}^{2N \times 2N}}$
    'LL': ('L','L'),
    # $\mathcolor{CommentColor}{\mathbf{M_{\mathcal{LC}}} \in \mathbb{B}^{2N \times M}}$
    'LC': ('L','C')
  },
  {
    # $\mathcolor{CommentColor}{\mu_{\mathcal{L} \rightarrow \mathcal{C}} : \tau_{\mathcal{L}} \rightarrow \tau_{\mathcal{C}}}$
    'L_msg_C': ('L','C'),
    # $\mathcolor{CommentColor}{\mu_{\mathcal{C} \rightarrow \mathcal{L}} : \tau_{\mathcal{C}} \rightarrow \tau_{\mathcal{L}}}$
    'C_msg_L': ('C','L')
  },
  {
    # $\mathcolor{CommentColor}{\mathbf{L}^{(t+1)}, \mathbf{L}_h^{(t+1)} \leftarrow \phi_{\mathcal{L}}(\mathbf{L}_h^{(t)},}$
    #       $\mathcolor{CommentColor}{\mathbf{M_{\mathcal{LC}}}^{\intercal} \times \mu_{\mathcal{L} \rightarrow \mathcal{C}}(\mathbf{C}^{(t)}), \mathbf{M_{\mathcal{LL}}} \times \mathbf{L}^{(t)} )}$
    'L': [
      {
        'mat': 'LC',
        'msg': 'L_msg_V',
        'transpose?': True,
        'var': 'L'
      },
      {
        'mat': 'LL',
        'var': 'L'
      }
    ],
    # $\mathcolor{CommentColor}{\mathbf{C}^{(t+1)}, \mathbf{C}_h^{(t+1)} \leftarrow \phi_{\mathcal{C}}(\mathbf{C}_h^{(t)}, \mathbf{M_{\mathcal{LC}}} \times \mu_{\mathcal{L} \rightarrow \mathcal{C}}(\mathbf{L}^{(t)}))}$
    'C': [
      {
        'mat': 'LC',
        'msg': 'L_msg_C',
        'var': 'L'
      }
    ]
  }
)
\end{lstlisting}
\end{minipage}
\end{figure}

\subsection{Solving the decision TSP} 

The TGN formalisation can be instantiated into the kernel of a end-to-end differentiable model to solve the decision version of the Traveling Salesperson Problem (TSP), although the corresponding block could conceivably be used to learn any graph problem with weighted edges (given an appropriate loss function). \cite{DBLP:journals/corr/abs-1809-02721} were able to train a GNN model with up to $80\%$ test accuracy on the decision problem (i.e. ``does graph $\mathcal{G} = (\mathcal{V}, \mathcal{E})$ admit a solution with cost no greater than $C$?'') by feeding it with pairs of complementary decision instances $X^{-} = (\mathcal{G}, 0.98 C^{*})$ and $X^{+} = (\mathcal{G}, 1.02 C^{*})$ where $\mathcal{G}$ is a random euclidean graph with optimal TSP cost $C^{*}$. The ground truth answers for the decision problem are \textbf{NO} and \textbf{YES} respectively for $X^{-}$ and $X^{+}$, forcing the model to learn to solve the problem within a $2\%$ relative deviation from the optimal cost.

Because edges are labeled with numerical information (i.e. their weights), it is convenient to assign embeddings to edges and nodes alike, and initialise edge embeddings with their corresponding weights. Then a TGN model can be instantiated in which nodes send messages to their incoming and outcoming edges, and edges send messages to their source and target nodes over many iterations of message passing. This model can be instantiated into the proposed TGN formalisation by defining an adjacency matrix ${\mathbf{M}_{\mathcal{E} \mathcal{V}} \in \mathbb{B}^{|\mathcal{E}| \times |\mathcal{V}|}}$ between edges and vertices (i.e. mapping each edge to its source and target vertices) and having the update step for vertex and edge embeddings be defined by:

\begin{equation}
\begin{aligned}
    \mathbf{V}^{(t+1)}, \mathbf{V}_h^{(t+1)} \leftarrow \phi_{\mathcal{V}}(\mathbf{V}_h^{(t)}, \mathbf{M_{\mathcal{E} \mathcal{V}}}^{\intercal} \times \mu_{\mathcal{E} \rightarrow \mathcal{V}}(\mathbf{E}^{(t)})) \\
    \mathbf{E}^{(t+1)}, \mathbf{E}_h^{(t+1)} \leftarrow \phi_{\mathcal{E}}(\mathbf{E}_h^{(t)}, \mathbf{M_{\mathcal{E} \mathcal{V}}} \times \mu_{\mathcal{V} \rightarrow \mathcal{E}}(\mathbf{V}^{(t)}))
\end{aligned}
\end{equation}

\begin{figure}[h]
    \centering
    \includegraphics[width=\linewidth]{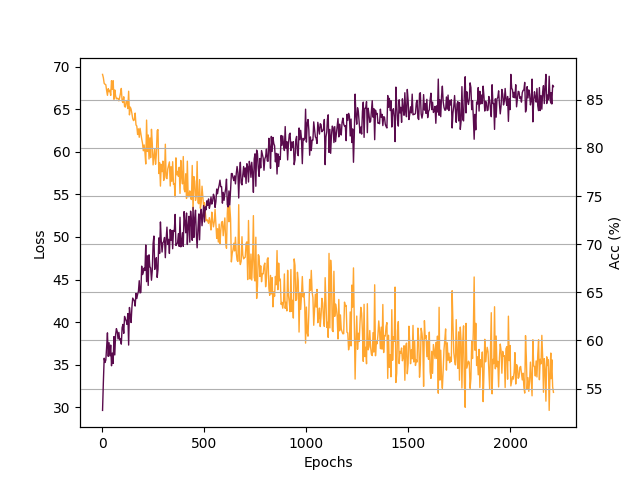}
    \caption{Results obtained with the model proposed \cite{DBLP:journals/corr/abs-1809-02721} implemented in the Typed Graph Networks library (see Code Listing \ref{TGN-TSP}).}
    \label{fig:TSP-training}
\end{figure}

\begin{figure}[h]
\begin{minipage}{\linewidth}
\begin{lstlisting}[language=Python,mathescape=true, caption={TGN kernel of an end-to-end differentiable model to solve the decision version of the Traveling Salesperson Problem, implemented with our Python library. Lines of code are commented with the corresponding equations in the proposed formalisation for TGNs in Algorithm \ref{alg:TGN}.},label=TGN-TSP]
tgn = TGN(
  {
    # $\mathcolor{CommentColor}{\tau_{\mathcal{V}} = \mathbb{R}^{d_v}}$
    'V': dv,
    # $\mathcolor{CommentColor}{\tau_{\mathcal{E}} = \mathbb{R}^{d_e}}$
    'E': de
  },
  {
    # $\mathcolor{CommentColor}{\mathbf{M_{\mathcal{E} \mathcal{V}}} \in \mathbb{B}^{|\mathcal{E}| \times |\mathcal{V}|}}$
    'EV': ('E','V')
  },
  {
    # $\mathcolor{CommentColor}{\tau_{\mathcal{V} \rightarrow \mathcal{E}} : \tau_{\mathcal{V}} \rightarrow \tau_{\mathcal{E}}}$
    'V_msg_E': ('V','E'),
    # $\mathcolor{CommentColor}{\tau_{\mathcal{E} \rightarrow \mathcal{V}} : \tau_{\mathcal{E}} \rightarrow \tau_{\mathcal{V}}}$
    'E_msg_V': ('E','V')
  },
  {
    # $\mathcolor{CommentColor}{\mathbf{V}^{(t+1)}, \mathbf{V}_h^{(t+1)} \leftarrow \phi_{\mathcal{V}}(\mathbf{V}_h^{(t)}, \mathbf{M_{\mathcal{E} \mathcal{V}}}^{\intercal} \times \mu_{\mathcal{E} \rightarrow \mathcal{V}}(\mathbf{E}^{(t)}))}$
    'V': [
      {
        'mat': 'EV',
        'msg': 'E_msg_V',
        'transpose?': True,
        'var': 'E'
      }
    ],
    # $\mathcolor{CommentColor}{\mathbf{E}^{(t+1)}, \mathbf{E}_h^{(t+1)} \leftarrow \phi_{\mathcal{E}}(\mathbf{E}_h^{(t)}, \mathbf{M_{\mathcal{E} \mathcal{V}}} \times \mu_{\mathcal{V} \rightarrow \mathcal{E}}(\mathbf{V}^{(t)}))}$
    'E': [
      {
        'mat': 'EV',
        'msg': 'V_msg_E',
        'var': 'V'
      }
    ]
  }
)
\end{lstlisting}
\end{minipage}
\end{figure}

\subsection{Ranking graph vertices by their centralities}

In a recent paper, \cite{avelar2018} trained a GNN model to predict centrality comparison on graphs (i.e. ``given a graph $\mathcal{G} = (\mathcal{V}, \mathcal{E})$, a centrality measure $c : \mathcal{V} \rightarrow \mathbb{R}$ and two vertices $v_1, v_2 \in \mathcal{V}$, does $c(v_1) < c(v_2)$ hold?''). The authors were able to achieve $89\%$ test accuracy on a dataset composed by random instances with up to $128$ vertices. Additionally, they were able to distill the same refined vertex embeddings into multiple centrality predictions, each corresponding to a different centrality measure, by training multiple MLPs fed with these embeddings at the end of the computation pipeline (multitask learning). This scenario can exemplify one of the simplest TGN models possible, corresponding to an iteration of the original Graph Neural Network model \cite{scarselli2009graph} with a single vertex type.

In their paper, \cite{avelar2018} differentiates between a message received from a source and from a target vertex. The update function can be described by the following equation:

\begin{equation}
\begin{aligned}
    \mathbf{V}^{(t+1)}, \mathbf{V_h}^{(t+1)} \leftarrow \phi_{\mathcal{V}}(\mathbf{V_h}^{(t)}, \mathbf{M} \times {\mu^{S}}_{\mathcal{V} \rightarrow \mathcal{V}}(\mathbf{V}^{(t)}), \\ \mathbf{M}^{\intercal} \times {\mu^{T}}_{\mathcal{V} \rightarrow \mathcal{V}}(\mathbf{V}^{(t)}))
\end{aligned}
\end{equation}

\begin{figure}[h]
    \centering
    \includegraphics[width=\linewidth]{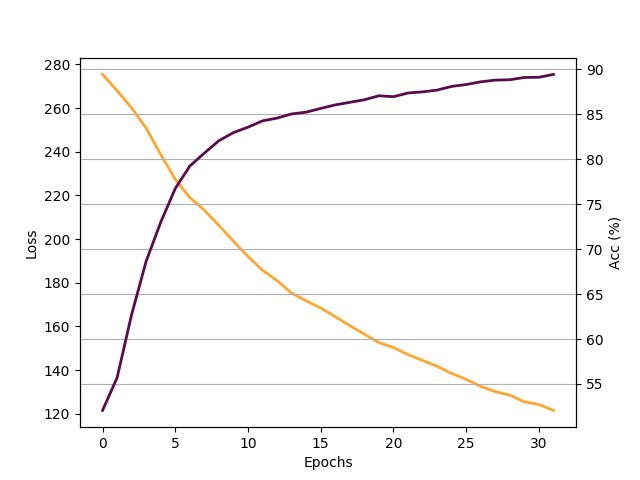}
    \caption{Results obtained with the model proposed in \cite{avelar2018} implemented with the Typed Graph Networks library (see Code Listing \ref{TGN-CENT}).}
    \label{fig:centrality-training}
\end{figure}

\begin{figure}[h]
\begin{minipage}{\linewidth}
\begin{lstlisting}[language=Python,mathescape=true, caption={TGN kernel of an end-to-end differentiable model to predict vertices ordering according to a given centrality, note that the resulting vertices embeddings should further be fed to $c$ MLPs in order to compute vertex-to-vertex comparisons},label=TGN-CENT]
tgn = GraphNN(
  {
    # $\mathcolor{CommentColor}{\tau_{\mathcal{V}} \in \mathbb{R}^{d}}$
    'V': d
  },
  {
    # $\mathcolor{CommentColor}{\mathbf{M} \in \mathbb{B}^{|\mathcal{V}| \times |\mathcal{V}|}}$
    'M': ('V','V')
  },
  {
    # $\mathcolor{CommentColor}{\mu^{S}_{\mathcal{V} \rightarrow \mathcal{C}} : \tau_{\mathcal{V}} \rightarrow \tau_{\mathcal{V}}}$
    'Vsource': ('V','V'),
    # $\mathcolor{CommentColor}{\mu^{T}_{\mathcal{V} \rightarrow \mathcal{C}} : \tau_{\mathcal{V}} \rightarrow \tau_{\mathcal{V}}}$
    'Vtarget': ('V','V')
  },
  {
    # $\mathcolor{CommentColor}{\mathbf{V}^{(t+1)}, \mathbf{V}_h^{(t+1)} \negthickspace \leftarrow \phi_{\mathcal{V}}(\mathbf{V}_h^{(t)}, \mathbf{M} \times {\mu^S}_{\mathcal{V} \rightarrow \mathcal{V}}(\mathbf{V}^{(t)}),}$ 
                   $\mathcolor{CommentColor}{\mathbf{M}^{\intercal} \negthickspace \times \negthickspace {\mu^T}_{\mathcal{C} \rightarrow \mathcal{V}}(\mathbf{V}^{(t)}))}$
    'V': [
      {
        'mat': 'M',
        'var': 'V',
        'msg': 'Vsource'
      },
      {
        'mat': 'M',
        'transpose?': True,
        'var': 'V',
        'msg': 'Vtarget'
      }
    ]
  },
)
\end{lstlisting}
\end{minipage}
\end{figure}

\subsection{Solving the Vertex k-Colorability Problem}\label{ssec:coloring-impl}

We aforementioned how the Graph Network model does not allow multiple embeddings corresponding to multiple global attributes for a graph (Section \ref{sec:form-tgn}). However this feature can be useful to solve a wide range of graph problems. We exemplify this issue here with the vertex coloring problem -- in which, given a graph $\mathcal{G} = (\mathcal{V},\mathcal{E})$ and an integer $k \in \mathbb{N}$, we must decide whether all the vertices $v \mathcal{V}$ can be partitioned into $k$ disjoint subsets (each representing a color) such that no two vertices $v_i, v_j$ from the same subset are connected by an edge $(i,j) \in \mathcal{E}$. In this context, it is useful to assign $k$ embeddings, one for each color, in addition to vertices embeddings. Color embeddings correspond to global attributes, as they should communicate with all vertex embeddings (i.e. each embedding is unconstrained, in principle, regarding which color it can assume). 

A TGN-based algorithm to tackle this problem could be defined as follows: initially, besides of the vertices adjacency matrix $\mathbf{M_{\mathcal{VV}}}$, we need to instantiate an adjacency matrix between vertices and colors $\mathbf{M_{\mathcal{VC}}}$, initialised with ones since a priori any color can be assigned to any vertex (this can be changed if one wishes to change the amount of initial information fed to the algorithm); then both vertices and colors embeddings are initialised - the later ones can be randomly initialised ($\sim \mathcal{U}[0,1)$) but ideally they could be placed equidistant over a hypersphere. Next, both vertices and colors embeddings should be updated throughout $t_{max}$ iterations according to the following set of equations:

\begin{equation}
\begin{aligned}
\mathbf{V}^{(t+1)}, \mathbf{V}_h^{(t+1)} \leftarrow \phi_{\mathcal{V}}(\mathbf{V}_h^{(t)}, \mathbf{M_{\mathcal{VV}}} \times (\mathbf{V}^{(t)}), \\
\mathbf{M_{\mathcal{VC}}} \times \mu_{\mathcal{C} \rightarrow \mathcal{V}}(\mathbf{C}^{(t)})) \\
\mathbf{C}^{(t+1)}, \mathbf{C}_h^{(t+1)} \leftarrow \phi_{\mathcal{C}}(\mathbf{C}_h^{(t)}, \mathbf{M_{\mathcal{VC}}}^{\intercal} \times \mu_{\mathcal{V} \rightarrow \mathcal{C}}(\mathbf{V}^{(t)}))
\end{aligned}
\end{equation}

Finally, each vertex embedding sums up to a final vote on whether the entire graph is $k$-colorable or not. At this point, one could also envision another possible architecture where color embeddings are also translated into votes and used, alone or combined with vertices embeddings, to decide the final answer.

\begin{figure}[h]
    \centering
    \includegraphics[width=\linewidth]{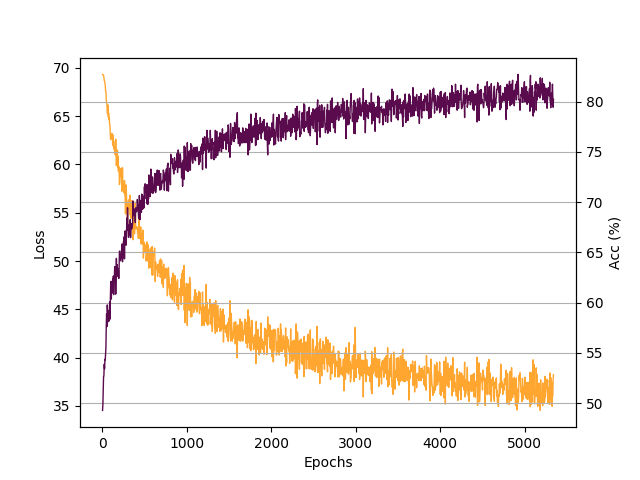}
    \caption{Results obtained with the implementation of a $k$-colorability solver in the Typed Graph Networks library (see Code Listing \ref{TGN-COL}).}
    \label{fig:kcolor-training}
\end{figure}

\begin{figure}[h]
\begin{minipage}{\linewidth}
\begin{lstlisting}[language=Python,mathescape=true, caption={TGN kernel of an end-to-end differentiable model to answer whether or not a given graph accepts a $k$ coloring, in this model only vertices embeddings are taken into account to compose the final answer},label=TGN-COL]
tgn = GraphNN(
  {
    # $\mathcolor{CommentColor}{\tau_{\mathcal{V}} = \mathbb{R}^{d_v}}$
    'V': dv,
    # $\mathcolor{CommentColor}{\tau_{\mathcal{C}} = \mathbb{R}^{d_c}}$
    'C': de 
  },
  {
    # $\mathcolor{CommentColor}{\mathbf{M} \in \mathbb{B}^{|\mathcal{V}| \times |\mathcal{V}|}}$
    'VV': ('V','V'),
    # $\mathcolor{CommentColor}{\mathbf{VC} \in \mathbbm{1}^{|\mathcal{V}| \times |\mathcal{C}|}}$
    'VC': ('V','C') 
  },
  {
    # $\mathcolor{CommentColor}{\mu_{\mathcal{V} \rightarrow \mathcal{C}} : \tau_{\mathcal{V}} \rightarrow \tau_{\mathcal{C}}}$
    'V_msg_C': ('V','C'),
    # $\mathcolor{CommentColor}{\mu_{\mathcal{C} \rightarrow \mathcal{V}} : \tau_{\mathcal{C}} \rightarrow \tau_{\mathcal{V}}}$
    'C_msg_V': ('C','V')
  },
  { 
    # $\mathcolor{CommentColor}{\mathbf{V}^{(t+1)}, \mathbf{V}_h^{(t+1)} \leftarrow \phi_{\mathcal{V}}(\mathbf{V}_h^{(t)}, \mathbf{M_{\mathcal{VV}}} \times \mathbf{V}^{(t)},}$
                $\mathcolor{CommentColor}{\mathbf{M_{\mathcal{VC}}} \times \mu_{\mathcal{C} \rightarrow \mathcal{V}}(\mathbf{C}^{(t)}))}$
    'V': [
      {
         'mat': 'M',
         'var': 'V'
      },
      {
         'mat': 'VC',
         'var': 'C',
         'msg': 'C_msg_V'
      }
    ],
    # $\mathcolor{CommentColor}{\mathbf{C}^{(t+1)}, \mathbf{C}_h^{(t+1)} \leftarrow \phi_{\mathcal{C}}(\mathbf{C}_h^{(t)}, \mathbf{M_{\mathcal{VC}}}^{\intercal} \times \mu_{\mathcal{V} \rightarrow \mathcal{C}}(\mathbf{V}^{(t)}))}$
    'C': [
      {
        'mat': 'VC',
        'msg': 'V_msg_C',
        'transpose?': True,
        'var': 'V'
      }
    ]
  },
)
\end{lstlisting}
\end{minipage}
\end{figure}

\section{Conclusion}
\label{sec:conc}

In the last few years, the deep learning community has experienced a boom in neural models engineered to learn on relational domains. Families of such models have been referred to using different names, such as Graph Neural Networks, Relational Networks and Message-passing Neural Networks. A recent paper by \cite{battaglia2018relational} provides a nice generalisation of many of these families into an unified formalisation which projects both nodes, edges and the entire graph into hyperdimensional real spaces. In this paper, we propose a related approach that benefits from diverting from a graph-centred ontology 
to a type-centred approach, as previously proposed by \cite{scarselli2009graph} (where the authors refer to types as ``kinds'' of vertices), which can easily capture higher level contexts such as hypergraphs. The Typed Graph Network formalisation, which partitions graph vertices into a number of \textbf{types}, allows one to define a Graph Neural Network model with a logical separation between different entities in the ontology of the problem one wants to learn. This is particularly useful for problems in which two or more global attributes must be considered -- for example, on a model for predicting $k$-colorability it is useful to have $k$ ``color'' embeddings each of which can be seen as a global attribute (i.e. each color communicates with all vertices in the graph). Apart from capturing a wider range of models, we argue that such a formalisation can also simplify their description when the problems under consideration does not necessarily require that edges and graphs have their own embeddings. As a companion to this paper, we provide a Python/Tensorflow library which allows one to easily compile TGNs into Tensorflow computation graphs by describing them in a manner consistent to the formalisation proposed.

We are hopeful that by thinking about graph neural networks in terms of types, these powerful techniques can be understood and effectively employed by a wider audience of deep learning researchers.

\subsection*{Acknowledgements}

We would like to thank Moshe Vardi for several suggestions and conversations that contributed to this research.
We would also like to thank NVIDIA for the GPU granted to our research group.

\bibliographystyle{IEEEtran}
\bibliography{tgn}

\end{document}